\documentclass[10pt,twocolumn,letterpaper]{article}

\usepackage{cvpr}              

\usepackage{graphicx}
\usepackage{amsmath}
\usepackage{amssymb}
\usepackage{booktabs}
\newcommand*{\thead}[1]{%
\multicolumn{1}{c}{\begin{tabular}{@{}c@{}}#1\end{tabular}}}
\usepackage{multirow}
\usepackage[para]{footmisc}

\makeatletter
\@namedef{ver@everyshi.sty}{}
\makeatother
\usepackage{pgfplots}
\usetikzlibrary{patterns}

\usepackage[pagebackref,breaklinks,colorlinks]{hyperref}

\usepackage[capitalize]{cleveref}
\crefname{section}{Sec.}{Secs.}
\Crefname{section}{Section}{Sections}
\Crefname{table}{Table}{Tables}
\crefname{table}{Table}{Tabs.}

\def\cvprPaperID{*****} 
\def\confName{CVPR}
\def\confYear{2022}

\begin{document}

\title{A Battle of Network Structures: An Empirical Study of CNN, Transformer, and MLP}

\author{

Yucheng Zhao\footnotemark[1] \footnotemark[2] $^{1}$ \qquad Guangting Wang\footnotemark[1] \footnotemark[2] $^{1}$ \qquad Chuanxin Tang\footnotemark[1] $^{2}$\qquad Chong Luo$^{2}$ \qquad Wenjun Zeng$^{2}$ \\ Zheng-Jun Zha $^{1}$\\
University of Science and Technology of China$^{1}$ \qquad Microsoft Research Asia$^{2}$ \\
{\tt\small \{lnc, flylight\}@mail.ustc.edu.cn \quad \{chutan, cluo, wezeng\}@microsoft.com\quad zhazj@ustc.edu.cn}

}
\maketitle

\renewcommand{\thefootnote}{\fnsymbol{footnote}}
\footnotetext[1]{Equal contribution.}
\footnotetext[2]{Interns at MSRA.}
\renewcommand{\thefootnote}{\arabic{footnote}}

\begin{abstract}

Convolutional neural networks (CNN) are the dominant deep neural network (DNN) architecture for computer vision. Recently, Transformer and multi-layer perceptron (MLP)-based models, such as Vision Transformer and MLP-Mixer, started to lead new trends as they showed promising results in the ImageNet classification task. In this paper, we conduct empirical studies on these DNN structures and try to understand their respective pros and cons. To ensure a fair comparison, we first develop a unified framework called SPACH which adopts separate modules for spatial and channel processing. Our experiments under the SPACH framework reveal that all structures can achieve competitive performance at a moderate scale. However, they demonstrate distinctive behaviors when the network size scales up. Based on our findings, we propose two hybrid models using convolution and Transformer modules. The resulting Hybrid-MS-S+ model achieves 83.9\% top-1 accuracy with 63M parameters and 12.3G FLOPS. It is already on par with the SOTA models with sophisticated designs. The code and models are publicly available at \url{https://github.com/microsoft/SPACH}.

\end{abstract}

\section{Introduction}

Convolutional neural networks (CNNs) have been dominating the computer vision (CV) field since the renaissance of deep neural networks (DNNs). They have demonstrated effectiveness in numerous vision tasks from image classification \cite{DBLP:conf/cvpr/HeZRS16}, object detection\cite{DBLP:journals/pami/RenHG017}, to pixel-based segmentation \cite{DBLP:conf/iccv/HeGDG17}. Remarkably, despite the huge success of Transformer structure \cite{DBLP:conf/nips/VaswaniSPUJGKP17} in natural language processing (NLP) \cite{DBLP:conf/naacl/DevlinCLT19}, the CV society still focuses on the CNN structure for quite some time. 

The transformer structure finally made its grand debut in CV last year. 
Vision Transformer (ViT) \cite{DBLP:conf/iclr/DosovitskiyB0WZ21} showed that a pure Transformer applied directly to a sequence of image patches can perform very well on image classification tasks, if the training dataset is sufficiently large. DeiT \cite{DBLP:conf/icml/TouvronCDMSJ21} further demonstrated that Transformer can be successfully trained on typical-scale dataset, such as ImageNet-1K \cite{DBLP:conf/cvpr/DengDSLL009}, with appropriate data augmentation and model regularization.

Interestingly, before the heat of Transformer dissipated, the structure of multi-layer perceptrons (MLPs) was revived by Tolstikhin et al. in a work called MLP-Mixer \cite{DBLP:journals/corr/abs-2105-01601}. MLP-Mixer is based exclusively on MLPs applied across spatial locations and feature channels. When trained on large datasets, MLP-Mixer attains competitive scores on image classification benchmarks. The success of MLP-Mixer suggests that neither convolution nor attention are necessary for good performance. It sparked further research on MLP as the authors wished \cite{DBLP:journals/corr/abs-2105-08050,DBLP:journals/corr/abs-2107-00645}. 

However, as the reported accuracy on image classification benchmarks continues to increase by new network designs from various camps, no conclusion can be made as which structure among CNN, Transformer, and MLP performs the best or is most suitable for vision tasks. This is partly due to the pursuit of high scores that leads to multifarious tricks and exhaustive parameter tuning. As a result, network structures cannot be fairly compared in a systematic way. The work presented in this paper fills this blank by conducting a series of controlled experiments over CNN, Transformer, and MLP in a unified framework.

\begin{figure*}[t]
  \centering
   \includegraphics[width=1\linewidth]{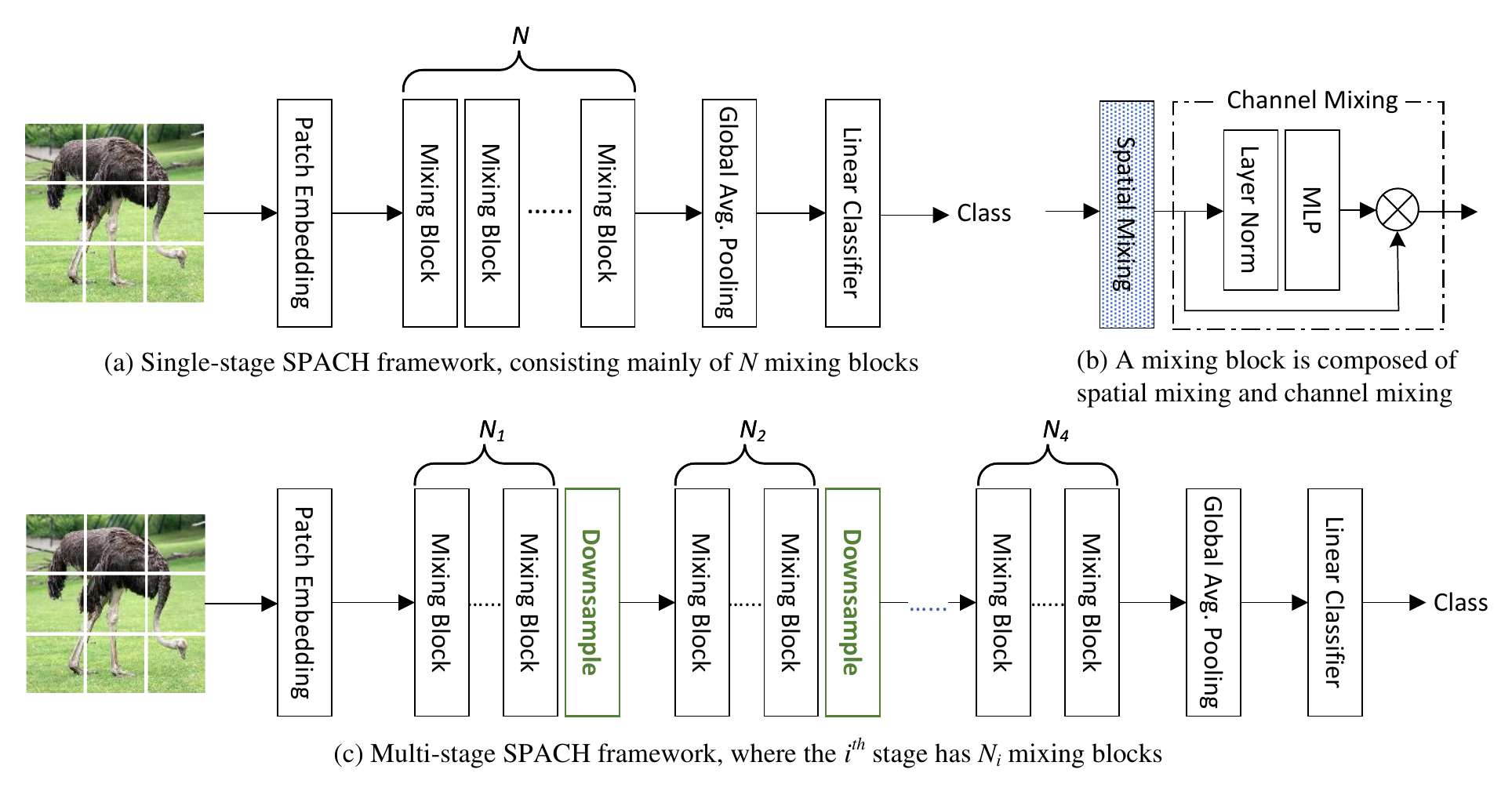}
   \caption{Illustration of the proposed experimental framework named SPACH.}
   \label{fig:SPACH}
\end{figure*}

We first develop a unified framework called SPACH as shown in Fig. \ref{fig:SPACH}. It is mostly adopted from current Transformer and MLP frameworks, since convolution can also fit into this framework and is in general robust to optimization. The SPACH framework contains a plug-and-play module called mixing block which could be implemented as convolution layers, Transformer layers, or MLP layers. Aside from the mixing block, other components in the framework are kept the same when we explore different structures. This is in stark contrast to previous work which compares different network structures in different frameworks that vary greatly in layer cascade, normalization, and other non-trivial implementation details. As a matter of fact, we found that these structure-free components play an important role in the final performance of the model, and this is commonly neglected in the literature. 

With this unified framework, we design a series of controlled experiments to compare the three network structures. The results show that all three network structures could perform well on the image classification task when pre-trained on ImageNet-1K. In addition, each individual structure has its distinctive properties leading different behaviors when the network size scales up. We also find several common design choices which contribute a lot to the performance of our SPACH framework. The detailed findings are listed in the following.

\begin{itemize}
    \item  Multi-stage design is standard in CNN models, but its effectiveness is largely overlooked in Transformer-based or MLP-based models. We find that the multi-stage framework consistently and notably outperforms the single-stage framework no matter which of the three network structures is chosen.
    
    \item Local modeling is efficient and crucial. With only light-weight depth-wise convolutions, the convolution model can achieve similar performance as a Transformer model in our SPACH framework. By adding a local modeling bypass in both MLP and Transformer structures, a significant performance boost is obtained with negligible parameters and FLOPs increase.
    
    \item MLP can achieve strong performance under small model sizes, but it suffers severely from over-fitting when the model size scales up. We believe that over-fitting is the main obstacle that prevents MLP from achieving SOTA performance.
    
    \item Convolution and Transformer are complementary in the sense that convolution structure has the best generalization capability while Transformer structure has the largest model capacity among the three structures. This suggests that convolution is still the best choice in designing lightweight models but designing large models should take Transformer into account.
    
\end{itemize}
    
Based on these findings, we propose two hybrid models of different scales which are built upon convolution and Transformer layers. Experimental results show that, when a sweet point between generalization capability and model capacity is reached, the performance of these straightforward hybrid models is already on par with SOTA models with sophisticated architecture designs.

\section{Background}

CNN and its variants have dominated the vision domain. During the evolution of CNN models, useful experience about the architecture design has been accumulated. Recently, two types of architectures, namely Transformer \cite{DBLP:conf/iclr/DosovitskiyB0WZ21} and MLP \cite{DBLP:journals/corr/abs-2105-01601}, begin to emerge in the vision domain and have shown performance similar to the well-optimized CNNs. These results kindle a spark towards building better vision models beyond CNNs.

\textbf{Convolution-based vision models}
Since the entrance of deep learning era pioneered by AlexNet \cite{DBLP:conf/nips/KrizhevskySH12}, the computer vision community has devoted enormous efforts to designing better vision backbones. In the past decade, most work focused on improving the design of CNN, and a series of networks, including VGG \cite{DBLP:journals/corr/SimonyanZ14a}, ResNet \cite{DBLP:conf/cvpr/HeZRS16}, SENet \cite{DBLP:conf/cvpr/HuSS18}, Xception \cite{DBLP:conf/cvpr/Chollet17}, MoblieNet\cite{DBLP:journals/corr/HowardZCKWWAA17,DBLP:conf/cvpr/SandlerHZZC18}, and EfficientNet \cite{DBLP:conf/icml/TanL19,DBLP:conf/icml/TanL21}, are designed. They achieve significant accuracy improvements in various vision tasks. 

A standard convolution layer learns filters in a 3D space, with two spatial dimensions and one channel dimension. Thus, the learning of spatial correlations and channel correlations are coupled inside a single convolution kernel. Differently, A depth-wise convolution layer only learns spatial correlations by moving the learning process of channel correlations to an additional 1x1 convolution. The fundamental hypothesis behind this design is that cross-channel correlations and spatial correlations are sufficiently decoupled that it is preferable not to map them jointly \cite{DBLP:conf/cvpr/Chollet17}. Recent work \cite{DBLP:conf/icml/TanL19,DBLP:conf/icml/TanL21} shows that depth-wise convolution can achieve both high accuracy and good efficiency, confirming this hypothesis to some extent. In addition, the idea of decoupling spatial and channel correlations is adopted in the vision Transformer. Therefore, this paper employs the spatial-channel decoupling idea in our framework design.

\textbf{Transformer-based vision models.} 
With the success of Transformer in natural language processing (NLP) \cite{DBLP:conf/nips/VaswaniSPUJGKP17,DBLP:conf/naacl/DevlinCLT19}, many researchers start to explore the use of Transformer as a stand-alone architecture for vision tasks. They are facing two main challenges. First, Transformer operates over a group of tokens, but no natural tokens, similar to the words in natural language, exist in an image. Second, images have a strong local structure while the Transformer structure treats all tokens equally and ignores locality. The pioneering work ViT \cite{DBLP:conf/iclr/DosovitskiyB0WZ21} solved the first challenge by simply dividing an image into non-overlapping patches and treat each patch as a visual token. ViT also reveals that Transformer models trained on large-scale datasets could attain SOTA image recognition performance. However, when the training data is insufficient, ViT does not perform well due to the lack of inductive biases. DeiT \cite{DBLP:conf/icml/TouvronCDMSJ21} mitigates the problem by introducing a regularization and augmentation pipeline on ImageNet-1K. 

Swin \cite{DBLP:journals/corr/abs-2103-14030} and Twins \cite{DBLP:journals/corr/abs-2104-13840} propose local ViT to address the second challenge. They adopt locally-grouped self-attention by computing the standard self-attention within non-overlapping windows. The local mechanism not only leads to performance improvement thanks to the reintroduction of locality, but also bring sufficient improvement on memory and computational efficiency. Thus, the pyramid structure becomes feasible again for vision Transformer.

There has been a blowout development in the design of Transformer-based vision models. Since this paper is not intended to review the progress of vision Transformer, we only briefly introduce some highly correlated Transformer models. CPVT \cite{chu2021conditional} and CvT \cite{DBLP:journals/corr/abs-2103-15808} introduce convolution into Transformer blocks, bringing the desired translation-invariance properties into ViT architecture. CaiT \cite{DBLP:journals/corr/abs-2103-17239} introduces a LayerScale approach to empower effective training of deeper ViT network. It is also discovered that some class-attention layers built on top of ViT network offer more effective processing than the class embedding. LV-ViT \cite{jiang2021all} proposes a bag of training techniques to build a strong baseline for vision Transformer. LeViT \cite{DBLP:journals/corr/abs-2104-01136} proposes a hybrid neural network for fast image classification inference.

\textbf{MLP-based vision models.} 
Although MLP is not a new concept for the computer vision community, the recent progress on MLP-based visual models surprisingly demonstrates, both conceptually and technically, that simple architecture can achieve competitive performance with CNN or Transformer \cite{DBLP:journals/corr/abs-2105-01601}. The pioneering work MLP-Mixer proposed a Mixer architecture using channel-mixing MLPs and channel-mixing MLPs to communicate between different channels and spatial locations (tokens), respectively. It achieves promising results when trained on a large-scale dataset (i.e., JFT\cite{DBLP:conf/iccv/SunSSG17}). ResMLP \cite{DBLP:journals/corr/abs-2105-03404} built a similar MLP-based model with a deeper architecture. ResMLP does not need large-scale datasets and it achieves comparable accuracy/complexity trade-offs on ImageNet-1K with Transformer-based models. FF \cite{DBLP:journals/corr/abs-2105-02723} showed that simply replacing the attention layer in ViT with an MLP layer applied over the patch dimension could achieve moderate performance on ImageNet classification. gMLP \cite{DBLP:journals/corr/abs-2105-08050} proposed a gating mechanism on MLP and suggested that self-attention is not a necessary ingredient for scaling up machine learning models.

\section{A Unified Experimental Framework}

In order to fairly compare the three network structures, we are in need of a unified framework that excludes other performance-affecting factors. Since recent MLP-based networks have already shared a similar framework as Transformer-based networks, we build the unified experimental framework based on them and try to include CNN-based network in this framework as well.

\begin{table*}
  \centering
  \begin{tabular}{@{}ccccc@{}}
    \toprule
    Model          & SPACH-XXS & SPACH-XS & SPACH-S \\ 
    \midrule
    Conv            & $C=384,R=2.0,N=12$ & $C=384,R=2.0,N=24$ & $C=512,R=3.0,N=24$\\
    \midrule
    Transformer     & $C=192,R=2.0,N=12$ & $C=384,R=2.0,N=12$ & $C=512,R=3.0,N=12$\\
    \midrule
    MLP             & $C=384,R=2.0,N=12$ & $C=384,R=2.0,N=24$ & $C=512,R=3.0,N=24$\\
    \midrule
    Conv-MS         & \thead{$C=64,R=2.0$\\$N_s=\{2,2,6,2\}$} & \thead{$C=96,R=2.0$\\$N_s=\{3,4,12,3\}$} & \thead{$C=128,R=3.0$\\$N_s=\{3,4,12,3\}$}\\
    \midrule
    Transformer-MS  & \thead{$C=32,R=2.0$\\$N_s=\{2,2,6,2\}$} & \thead{$C=64,R=2.0$\\$N_s=\{3,4,12,3\}$} & \thead{$C=96,R=3.0$\\$N_s=\{3,4,12,3\}$}\\
    \midrule
    MLP-MS          & \thead{$C=64,R=2.0$\\$N_s=\{2,2,6,2\}$} & \thead{$C=96,R=2.0$\\$N_s=\{3,4,12,3\}$} & \thead{$C=128,R=3.0$\\$N_s=\{3,4,12,3\}$}\\
    
    \bottomrule
  \end{tabular}
  \caption{SPACH and SPACH-MS model variants. $C$: feature dimension, $R$: expansion ratio of MLP in $\mathcal{F}_c$, $N$: number of mixing blocks of SPACH, $N_s$: number of mixing blocks in the $i_{th}$ stage of SPACH-MS.}
  \label{tab:configuration}
\end{table*}

\subsection{Overview of the SPACH Framework}

We build our experimental framework with reference to ViT \cite{DBLP:conf/iclr/DosovitskiyB0WZ21} and MLP-Mixer \cite{DBLP:journals/corr/abs-2105-01601}. Fig. \ref{fig:SPACH}(a) shows the single-stage version of the SPACH framework, which is used for our empirical study. The architecture is very simple and consists mainly of a cascade of mixing blocks, plus some necessary auxiliary modules, such as patch embedding, global average pooling, and a linear classifier. Fig. \ref{fig:SPACH}(b) shows the details of the mixing block. Note that the spatial mixing and channel mixing are performed in consecutive steps. The name SPACH for our framework is coined to emphasize the serial structure of SPAtial and CHannel processing. 

We also enable a multi-stage variation, referred to as SPACH-MS, as shown in Fig. \ref{fig:SPACH}(c). Multi-stage is an important mechanism in CNN-based networks to improve the performance. Unlike the single-stage SPACH, which processes the image in a low resolution by down-sampling the image by a large factor at the input, SPACH-MS is designed to keep a high-resolution in the initial stage of the framework and progressively perform down-sampling. Specifically, our SPACH-MS contains four stages with down-sample ratios of 4, 8, 16, and 32, respectively. Each stage contains $N_s$ mixing blocks, where $s$ is the stage index. Due to the extremely high computational cost of Transformer and MLP on high-resolution feature maps, we implement the mixing blocks in the first stage with convolutions only. The feature dimension within a stage remains constant, and will be multiplied with a factor of 2 after down-sampling.

Let $I\in \mathbb{R}^{3\times h\times w}$ denotes an input image, where 3 is the RGB channels and $H\times W$ is the spatial dimensions. Our SPACH framework first passes the input image through a Patch Embedding layer, which is the same as the one in ViT, to convert $I$ into patch embeddings $X_p\in \mathbb{R}^{C\times \frac{h}{p} \times \frac{w}{p}}$. Here $p$ denotes patch size, which is 16 in the single-stage implementation and 4 in the multi-stage implementation. 
After the cascades of mixing blocks, a classification head implemented by a linear layer is used for the supervised pre-training.

We list the hyper-parameters used in different model configurations in Table \ref{tab:configuration}. Three model size for each variations of SPACH are designed, namely SPACH-XXS, SPACH-XS and SPACH-S, by controlling the number of blocks, the number of channels, and the expansion ratio of channel mixing MLP $\mathcal{F}_c$. The model size, theoretical computational complexity (FLOPS), and empirical throughput are presented in Section \ref{sec:exp}. We measure the throughput using one P100 GPU.

\subsection{Mixing Block Design} \label{sec:3.1}

\begin{figure}[t]
  \centering
   \includegraphics[width=1.0\linewidth]{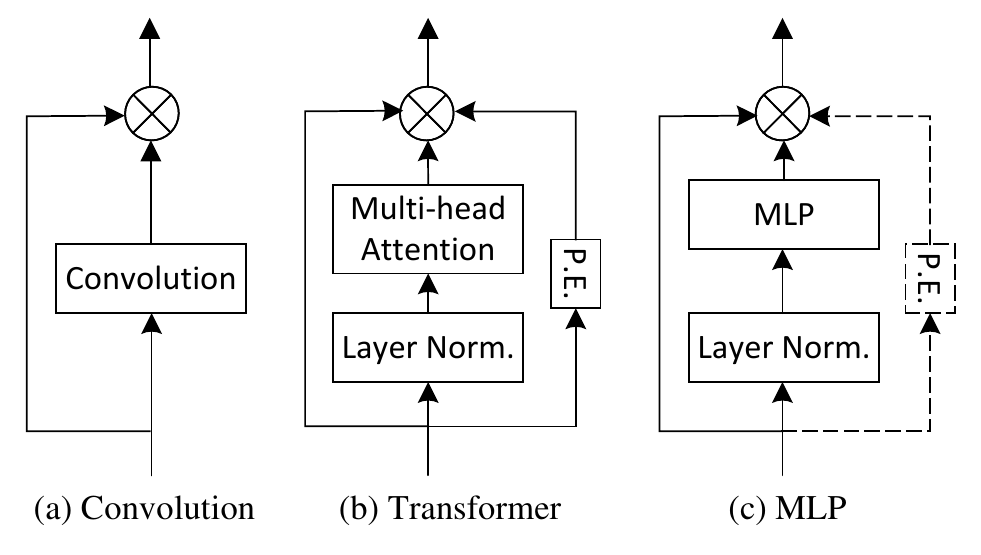}

   \caption{Three implementations of the spatial mixing module using convolution, Transformer, and MLP, respectively. P.E. denotes positional encoding, implemented by convolution in SPACH.}
   \label{fig:mixing block}
\end{figure}

Mixing blocks are key components in the SPACH framework. As shown in Fig. \ref{fig:SPACH}(b), for an input feature $X\in\mathbb{R}^{C\times H\times W}$, where $C$ and $H\times W$ denote channel and spatial dimensions, it is first processed by a spatial mixing function $\mathcal{F}_s$ and then by a channel mixing function $\mathcal{F}_c$. $\mathcal{F}_s$ focuses on aggregating context information from different spatial locations while $\mathcal{F}_c$ focuses on channel information fusion. Denoting the output as $Y$, we can formulate a mixing block as:
\begin{equation}
    Y = \mathcal{F}_s(\mathcal{F}_c(X)).
\end{equation}

Following ViT \cite{DBLP:conf/iclr/DosovitskiyB0WZ21}, we use an MLP with appropriate normalization and residual connection to implement $\mathcal{F}_c$. The MLP here can be also viewed as a 1x1 convolution (also known as point-wise convolution \cite{DBLP:conf/cvpr/Chollet17}) which is a special case of regular convolution. Note that $\mathcal{F}_c$ only performs channel fusion and does not explore any spatial context.

The spatial mixing function $\mathcal{F}_s$ is the key to implement different architectures. As shown in Fig. \ref{fig:mixing block}, we implement three structures using convolution, self-attention, and MLP. The common components include normalization and residual connection.
Specifically, the convolution structure is implemented by a 3x3 depth-wise convolution, as channel mixing will be handled separately in subsequent steps. For the Transformer structure, there is a positional embedding module in the original design. But recent research suggests that absolute positional embedding breaks translation variance, which is not suitable for images. In view of this and inspired by recent vision transformer design \cite{chu2021conditional,DBLP:journals/corr/abs-2103-15808}, we introduce a convolutional positional encoding (CPE) as a bypass in each spatial mixing module. The CPE module has negligible parameters and FLOPs. 
For MLP-based network, the pioneering work MLP-Mixer does not use any positional embedding, but we empirically find that adding the very lightweight CPE significantly improves the model performance, so we use the same treatment for MLP as for Transformer.  

The three implementations of $\mathcal{F}_s$ have distinctive properties as listed in Table \ref{tab:properites}. First, the convolution structure only involves local connections so that it is computational efficient. Second, the self-attention structure uses dynamic weight for each input instance so that model capacity is increased. Moreover, it has a global receptive field, which enables information to flow freely across different positions \cite{DBLP:conf/nips/VaswaniSPUJGKP17}. Third, MLP structure has a global receptive field just as the self-attention structure, but it does not use dynamic weight. In summary, these three properties seen in different architectures are all desirable and may have positive influence on the model performance or efficiency. We can find convolution and self-attention have complementary properties thus there is potential to build hybrid model to combine all desirable properties. Besides, MLP structure seems to be inferior to self-attention in this analysis.

\begin{table}
  \centering
  \begin{tabular}{@{}cccc@{}}
    \toprule
    Properties                          & Convolution & Self-Attention & MLP \\
    \midrule
    \thead{Sparse \\ Connectivity}      & \checkmark  &                &     \\
    \midrule
    \thead{Dynamic \\ Weight}           &             & \checkmark     &     \\
    \midrule
    \thead{Global \\ Receptive Field}   &             & \checkmark     & \checkmark \\
    \bottomrule
  \end{tabular}
  \caption{Three desired properties in network design are seen in different network structures.}
  \label{tab:properites}
\end{table}

\section{Empirical Studies on Mixing Blocks} \label{sec:exp}

In this section, we design a series of controlled experiments to compare the three network structures. We first introduce the experimental settings in Section \ref{subsec:settings}, and then present our main findings in Section \ref{subsec:exp1}, \ref{subsec:exp2}, \ref{subsec:exp3}, and \ref{subsec:exp4}.

\subsection{Datasets and Training Pipelines} \label{subsec:settings}

We conduct experiments on ImageNet-1K (IN-1K) \cite{DBLP:conf/cvpr/DengDSLL009} image classification which has 1k classes. The training set has 1.28M images while the validation set has 50k images. The Top-1 accuracy on a single crop is reported. Unless otherwise indicated, we use the input resolution of 224x224. Most of our training settings are inherited from DeiT \cite{DBLP:conf/icml/TouvronCDMSJ21}. We employ an AdamW \cite{DBLP:journals/corr/abs-1711-05101} optimizer for 300 epochs with a cosine decay learning rate scheduler and 20 epochs of linear warm-up. The weight decay is 0.05, and the initial learning rate is $0.005\times \frac{\text{batchsize}}{512}$. 8 GPUs with mini-batch 128 per GPU are used in training, resulting a total batch-size of 1024. We use exactly the same data augmentation and regularization configurations as DeiT, including Rand-Augment \cite{DBLP:conf/nips/CubukZS020}, random erasing \cite{DBLP:conf/aaai/Zhong0KL020}, Mixup \cite{DBLP:conf/iclr/ZhangCDL18}, CutMix \cite{DBLP:conf/iccv/YunHCOYC19}, stochastic depth \cite{DBLP:conf/eccv/HuangSLSW16}, and repeated augmentation \cite{DBLP:journals/corr/abs-1902-05509,DBLP:conf/cvpr/HofferBHGHS20}. We use the same training pipeline for all comparing models. And the implementation is built upon PyTorch \cite{DBLP:conf/nips/PaszkeGMLBCKLGA19} and timm library \cite{rw2019timm}.

\begin{table*}
  \centering
  \setlength\tabcolsep{3pt}
  \begin{tabular}{@{}c|c|ccc|c|ccc|c }
    \toprule
    \multirow{3}{*}{Network Scale}  & \multirow{3}{*}{Model}    & \multicolumn{4}{c}{Single-Stage}    \vline & \multicolumn{4}{c}{Multi-Stage}     \\ \cline{3-10}
    & & \multirow{2}{*}{\#param.}  & \multirow{2}{*}{FLOPs} & Throughput & IN-1K & \multirow{2}{*}{\#param.}  & \multirow{2}{*}{FLOPs} & Throughput & IN-1K  \\
    & & & & (image/s)  & Top-1 acc. & & & (image/s)  & Top-1 acc. \\
    \hline
    \multirow{3}{*}{XXS} 
    & Conv  & 8M    & 1.4G  & 1513  & 72.1  & 5M    & 0.7G \textcolor{red}{(-0.7G)}  & 1576  &73.3 \textcolor{red}{(+1.2)} \\
    & Trans     & 4M    & 0.9G     & 1202   & 68.0     & 2M     & 0.5G \textcolor{red}{(-0.4G)}   & 1558  & 65.4 \textcolor{green}{(-2.6)} \\
    & MLP   & 9M    & 1.8G  & 980   & 74.1  & 6M    & 0.9G \textcolor{red}{(-0.9G)}  & 1202    & 74.9 \textcolor{red}{(+0.8)}  \\
    \hline
    \multirow{3}{*}{XS} 
    &Conv   & 15M   & 2.8G    & 770     & 77.3    & 17M          & 2.8G \textcolor{red}{(-0.0G)}      & 602   & 80.1 \textcolor{red}{(+2.8)} \\
    &Trans  & 15M   & 3.1G    & 548     & 78.4   & 14M          & 3.1G \textcolor{red}{(-0.0G)}      & 441    & 80.1 \textcolor{red}{(+1.7)} \\
    &MLP    & 17M   & 3.5G    & 503     & 78.5   & 19M          & 3.4G \textcolor{red}{(-0.1G)}  & 438    & 80.7 \textcolor{red}{(+2.2)} \\
    \hline
    \multirow{3}{*}{S} 
    &Conv   & 39M          & 7.4G    & 374  & 80.1  & 44M          & 7.2G \textcolor{red}{(-0.2G)}      & 328   & 81.6 \textcolor{red}{(+1.5)}  \\
    &Trans  & 33M          & 6.7G    & 328  & 81.7  & 40M          & 7.6G \textcolor{green}{(+0.9G)}      & 246   & 82.9 \textcolor{red}{(+1.2)}  \\
    &MLP    & 41M          & 8.7G    & 272  & 78.6  & 46M          & 8.2G \textcolor{red}{(-0.5G)}      & 254     & 82.1 \textcolor{red}{(+3.5)} \\
    \bottomrule
  \end{tabular}
  \caption{Model performance of SPACH and SPACH-MS at three network scales.}
  \label{tab:exp1.1}
\end{table*}

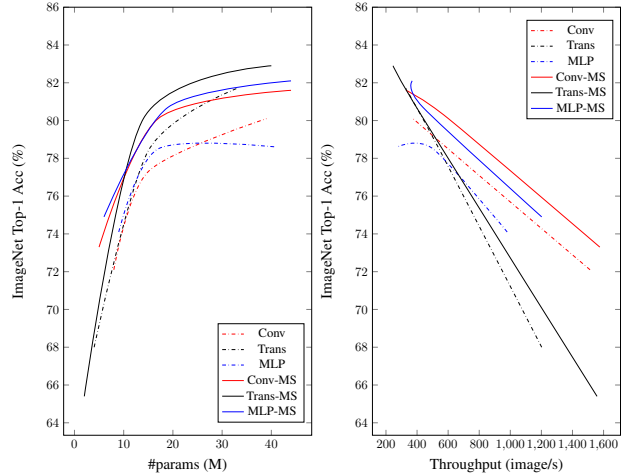
\begin{figure}[t]
    \centering
    \begin{subfigure}[b]{0.22\textwidth}
    \begin{tikzpicture}[scale=0.48]
    \begin{axis}[
    xlabel={\#params (M)},
    ylabel={ImageNet Top-1 Acc (\%)},
	ymax=86,
    height=3.5\linewidth,
    width=2.2\textwidth,
    legend pos=south east,
    legend style={nodes={scale=1.0, transform shape}},
    label style={font=\large}
    ]
    \addplot[dash dot, smooth,color=red] plot coordinates {
        (8, 72.1)
        (15, 77.3)
        (39, 80.1)
 
    };
    \addlegendentry{Conv }
    
    \addplot[dash dot, smooth,color=black] plot coordinates {
        (4, 68.0)
        (15, 78.4)
        (33, 81.7)
 
    };
    \addlegendentry{Trans }
    
    \addplot[dash dot, smooth,color=blue] plot coordinates {
        (9, 74.1)
        (17, 78.5)
        (41, 78.6)
 
    };
    \addlegendentry{MLP }
    
    \addplot[solid, smooth,color=red] plot coordinates {
        (5, 73.3)
        (17, 80.1)
        (44, 81.6)
 
    };
    \addlegendentry{Conv-MS }
    
    \addplot[solid, smooth,color=black] plot coordinates {
        (2, 65.4)
        (14, 80.1)
        (40, 82.9)
 
    };
    \addlegendentry{Trans-MS }
    
    \addplot[solid, smooth,color=blue] plot coordinates {
        (6, 74.9)
        (19, 80.7)
        (44, 82.1)
 
    };

    \addlegendentry{MLP-MS }
    
    \end{axis}
    \end{tikzpicture}
    \end{subfigure}    
    ~
    \begin{subfigure}[b]{0.22\textwidth}
    \begin{tikzpicture}[scale=0.48]
    \begin{axis}[
    xlabel={Throughput (image/s)},
    ylabel={ImageNet Top-1 Acc (\%)},
	ymax=86,
    height=3.5\linewidth,
    width=2.2\textwidth,
    legend pos=north east,
    legend style={nodes={scale=1.0, transform shape}},
    label style={font=\large}
    ]
    \addplot[dash dot, smooth,color=red] plot coordinates {
        (1513, 72.1)
        (770, 77.3)
        (374, 80.1)
 
    };
    \addlegendentry{Conv }
    
    \addplot[dash dot, smooth,color=black] plot coordinates {
        (1202, 68.0)
        (548, 78.4)
        (328, 81.7)
 
    };
    \addlegendentry{Trans }
    
    \addplot[dash dot, smooth,color=blue] plot coordinates {
        (980, 74.1)
        (503, 78.5)
        (272, 78.6)
 
    };
    \addlegendentry{MLP }
    
    \addplot[solid, smooth,color=red] plot coordinates {
        (1576, 73.3)
        (602, 80.1)
        (328, 81.6)
 
    };
    \addlegendentry{Conv-MS }
    
    \addplot[solid, smooth,color=black] plot coordinates {
        (1558, 65.4)
        (441, 80.1)
        (246, 82.9)
 
    };
    \addlegendentry{Trans-MS }
    
    \addplot[solid, smooth,color=blue] plot coordinates {
        (1202, 74.9)
        (438, 80.7)
        (369, 82.1)
 
    };
    \addlegendentry{MLP-MS }
    \end{axis}
    \end{tikzpicture}
    \end{subfigure}    
    \caption{The multi-stage models (named with -MS suffix) always achieve a better performance than their single-stage counterparts.}
    \label{fig:exp1.1}
    \end{figure}

\begin{table}
  \centering
  \setlength\tabcolsep{3pt}
  \begin{tabular}{l|ccc|c}
    \toprule
    \multirow{2}{*}{Model}       & \multirow{2}{*}{\#param.}  & \multirow{2}{*}{FLOPs} & throughput & IN-1K  \\
                                 &                            &                        & (image/s)  & Top-1 acc. \\
    \hline
    Trans-MS-S                   & 40M          & 7.6G      & 246                               & 82.9 \\
    $\text{Trans-MS-S}^{-}$                    & 40M          & 7.6G      & 259                               & 82.3 \\
    \hline
    MLP-MS-S                     & 46M          & 8.2G      & 254                               & 82.1  \\
    $\text{MLP-MS-S}^{-}$                    & 46M          & 8.2G      & 274                               & 80.1 \\
    \bottomrule
  \end{tabular}
  \caption{Both Transformer structure and MLP structure benefit from local modeling at a very small computational cost. The superscription - indicates without local modeling.}
  \label{tab:exp1.2}
\end{table}

\subsection{Multi-Stage is Superior to Single-Stage} \label{subsec:exp1}

Multi-stage design is standard in CNN models, but it is largely overlooked in Transformer-based or MLP-based models. Our first finding is that multi-stage design should always be adopted in vision models no matter which of the three network structures is chosen.

Table \ref{tab:exp1.1} compares the image classification performance between multi-stage framework and single-stage framework. For all three network scales and all three network structures, multi-stage framework consistently achieves better complexity-accuracy trade-off. For the sake of easy comparison, the changes of FLOPs and accuracy are highlighted in Table \ref{tab:exp1.1}. Most of the multi-stage models are designed to have slightly fewer computational costs, but they manage to achieve a higher accuracy than the corresponding single-stage models. An accuracy loss of 2.6 points is observed for the Transformer model at the XXS scale, but it is understandable as the multi-stage model happens to have only half of the parameters and FLOPs of the corresponding single-stage model.

In addition, Fig. \ref{fig:exp1.1} shows how the image classification accuracy changes with the size of model parameters and model throughput. Despite the different trends observed for different network structures, the multi-stage models always outperform their single-stage counterparts.

This finding is consistent with the results reported in recent work. Both Swin-Transformer \cite{DBLP:journals/corr/abs-2103-14030} and TWins \cite{DBLP:journals/corr/abs-2104-13840} adopt multi-stage framework and achieve a stronger performance than the single-stage framework DeiT \cite{DBLP:conf/icml/TouvronCDMSJ21}. 
Our empirical study suggests that the use of multi-stage framework can be an important reason.

\subsection{Local Modeling is Crucial} \label{subsec:exp2}

Although it has been pointed out in many previous work \cite{DBLP:journals/corr/abs-2103-15808,chu2021conditional,DBLP:journals/corr/abs-2104-05707,DBLP:conf/icml/dAscoliTLMBS21,DBLP:journals/corr/abs-2103-14030} that local modeling is crucial for vision models, we will show in this subsection how amazingly efficient local modeling could be. 

In our empirical study, the spatial mixing block of the convolution structure is implemented by a $3 \times 3$ depth-wise convolution, which is a typical local modeling operation. It is so light-weight that it only contributes to 0.3\% of the model parameter and 0.5\% of the FLOPs. However, as Table \ref{tab:exp1.1} and Fig. \ref{fig:exp1.1} show, this structure can achieve competitive performance when compared with the Transformer structure in the XXS and XS configurations.

It is due to the sheer efficiency of $3 \times 3$ depth-wise convolution that we propose to use it as a bypass in both MLP and Transformer structures. The increase of model parameters and inference FLOPs is almost negligible, but the locality of the models is greatly strengthened. In order to demonstrate how local modeling helps the performance of Transformer and MLP structures, we carry out an ablation study which removes this convolution bypass in the two structures. 

Table \ref{tab:exp1.2} shows the performance comparison between models with or without local modeling. The two models we pick are the top performers in Table \ref{tab:exp1.1} when multi-stage framework is used and network scale is S. We can clearly find that the convolution bypass only slightly decreases the throughput, but brings a notable accuracy increase to both models. Note that the convolution bypass is treated as convolutional positional embedding in Trans-MS-S, so we bring back the standard patch embedding as in ViT \cite{DBLP:conf/iclr/DosovitskiyB0WZ21} in $\text{Trans-MS-S}^{-}$. For $\text{MLP-MS-S}^{-}$, we follow the practice in MLP-Mixer and do not use any positional embedding. 
This experiment confirms the importance of local modeling and suggests the use of $3 \times 3$ depth-wise convolution as a bypass for any designed network structures.

\subsection{A Detailed Analysis of MLP} \label{subsec:exp3}

\begin{table}
  \centering
  \setlength\tabcolsep{3pt}
  \begin{tabular}{c|ccc|c}
    \toprule
    \multirow{2}{*}{Model}       & \multirow{2}{*}{\#param.}  & \multirow{2}{*}{FLOPs} & throughput & IN-1K  \\
                                 &                            &                        & (image/s)  & Top-1 acc. \\
    \hline
    MLP-S       & 41M          & 8.7G     & 272                           & 78.6 \\
    +Shared     & 39M          & 8.7G     & 274                           & 80.2 \\
    \hline
    MLP-MS-S    & 46M          & 8.2G     & 254                           & 82.1 \\
    +Shared     & 45M          & 8.2G     & 244                           & 82.5 \\

    \bottomrule
  \end{tabular}
  \caption{The performance of MLP models are greatly boosted when weight sharing is adopted to alleviate over-fitting. }
  \label{tab:exp3.2}
\end{table}

Due to the excessive number of parameters, MLP models suffer severely from over-fitting. We believe that over-fitting is the main obstacle for MLP to achieve SOTA performance. In this part, we discuss two mechanisms which can potentially alleviate this problem. 

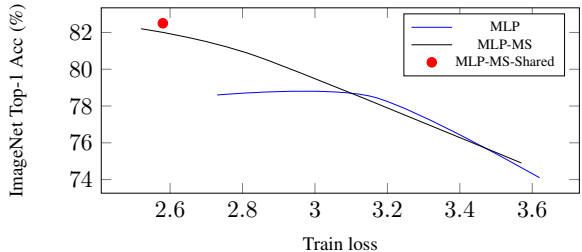
\begin{figure}[t]
    \centering
    \begin{subfigure}[b]{0.45\textwidth}
    \begin{tikzpicture}[scale=0.9]
    \begin{axis}[
    xlabel={Train loss},
    ylabel={ImageNet Top-1 Acc (\%)},
    height=0.55\linewidth,
    width=1.1\columnwidth,
    legend pos=north east,
    legend style={nodes={scale=0.6, transform shape}},
    label style={font=\footnotesize}
    ]

    \addplot[solid, smooth,color=blue] plot coordinates {

        (3.62, 74.1)
        (3.16, 78.5)
        (2.73, 78.6)
 
    };
    \addlegendentry{MLP }
    
        \addplot[solid, smooth,color=black] plot coordinates {

        (3.57, 74.9)
        (2.84, 80.7)
        (2.52, 82.2)
 
    };
    \addlegendentry{MLP-MS }
    
    \addplot[only marks, mark=*, color=red] plot coordinates {
        (2.58, 82.5)
    };
    \addlegendentry{MLP-MS-Shared }

    \end{axis}
    \end{tikzpicture}
    \end{subfigure}

    \caption{Illustration of the over-fitting problem in MLP-based models. Both multi-stage framework and weight sharing alleviate the problem.}
    \label{fig:exp3.1}
    \end{figure}

One is the use of multi-stage framework. We have already shown in  Table \ref{tab:exp1.1} that multi-stage framework brings gain. Such gain is even more prominent for larger MLP models. In particular, the MLP-MS-S models achieves 2.6 accuracy gain over the single-stage model MLP-S. We believe this owes to the strong generalization capability of the multi-stage framework. 
Fig. \ref{fig:exp3.1} shows how the test accuracy increases with the decrease of training loss. Over-fitting can be observed when the test accuracy starts to flatten. 
These results also lead to a very promising baseline for MLP-based models. Without bells and whistles, MLP-MS-S model achieves 82.1\% ImageNet Top-1 accuracy, which is 5.7 points higher than the best results reported by MLP-Mixer \cite{DBLP:journals/corr/abs-2105-01601} when ImageNet-1K is used as training data.

The other mechanism is parameter reduction through weight sharing. We apply weight-sharing on the spatial mixing function $\mathcal{F}_s$. For the single-stage model, all $N$ mixing blocks use the same $\mathcal{F}_s$, while for the multi-stage model, each stage use the same same $\mathcal{F}_s$ for its $N_s$ mixing blocks. We present the results of S models in Table \ref{tab:exp3.2}. We can find that the shared-weight variants, denoted by "+Shared", achieve higher accuracy with almost the same model size and computation cost. Although they are still inferior to Transformer models, the performance is on par with or even better than convolution models. 
Fig. \ref{fig:exp3.1} confirms that using shared weights in the MLP-MS model further delays the appearance of over-fitting signs. Therefore, we conclude that MLP-based models remain competitive if they could solve or alleviate the over-fitting problem. 

\subsection{Convolution and Transformer are Complementary} \label{subsec:exp4}

We find that convolution and Transformer are complementary in the sense that convolution structure has the best generalization capability while Transformer structure has the largest model capacity among the three structures we investigated. 

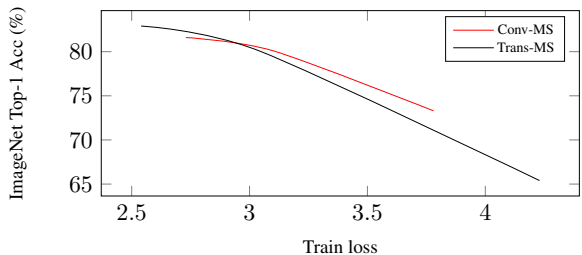
\begin{figure}[t]
    \centering
    
    \begin{subfigure}[b]{0.45\textwidth}
    \begin{tikzpicture}[scale=0.9]
    \begin{axis}[
    xlabel={Train loss},
    ylabel={ImageNet Top-1 Acc (\%)},
    height=0.55\linewidth,
    width=1.1\columnwidth,
    legend pos=north east,
    legend style={nodes={scale=0.6, transform shape}},
    label style={font=\footnotesize}
    ]

    \addplot[solid, smooth,color=red] plot coordinates {
        (3.78, 73.3)
        (3.10, 80.1)
        (2.73, 81.6)
 
    };
    \addlegendentry{Conv-MS }
    
    \addplot[solid, smooth,color=black] plot coordinates {
        (4.23, 65.4)
        (3.04, 80.1)
        (2.54, 82.9)
 
    };
    \addlegendentry{Trans-MS }

    \end{axis}
    \end{tikzpicture}
    \end{subfigure}    
    
    \caption{Conv-MS has a better generalization capability than Trans-MS as it achieves a higher test accuracy at the same training loss before the model saturates.}
    \label{fig:exp3.2}
    \end{figure}

Fig. \ref{fig:exp3.2} shows that, before the performance of Conv-MS saturates, it achieves a higher test accuracy than Trans-MS at the same training loss. This shows that convolution models generalize better than Transformer models. In particular, when the training loss is relatively large, the convolution models show great superiority against Transformer models. This suggests that convolution is still the best choice in designing lightweight vision models.

On the other hand, both Fig. \ref{fig:exp1.1} and Fig. \ref{fig:exp3.2} show that Transformer models achieve higher accuracy than the other two structures when we increase the model size and allow for higher computational cost. 
Recall that we have discussed three properties of network architectures in Section \ref{sec:3.1}. It is now clear that the sparse connectivity helps to increase generalization capability, while dynamic weight and global receptive field help to increase model capacity.

\section{Hybrid Models}

As discussed in Section \ref{sec:3.1} and \ref{subsec:exp3}, convolution and Transformer structures have complementary characteristics and have potential to be used in a single model. Based on this observation, we construct hybrid models at the XS and S scales based on these two structures. The procedure we used to construct hybrid models is rather simple. We take a multi-stage convolution-based model as the base model, and replace some selected layers with Transformer layers. Considering the local modeling capability of convolutions and global modeling capability of Transformers, we tend to do such replacement in later stages of the model. The details of layer selection in the two hybrid models are listed as follows.

\begin{itemize}
    \item Hybrid-MS-XS: It is based on Conv-MS-XS. The last ten layers in Stage 3 and the last two layers in Stage 4 are replaced by Transformer layers. Stage 1 and 2 remain unchanged.
    \item Hybrid-MS-S: It is based on Conv-MS-S. The last two layers in Stage 2, the last ten layers in Stage 3, and the last two layers in Stage 4 are replaced by Transformer layers. Stage 1 remains unchanged.  
\end{itemize}

In order to unleash the full potential of hybrid models, we further adopt the \textit{deep} patch embedding layer (PEL) implementation as suggested in LV-ViT \cite{jiang2021all}. Different from \textit{default} PEL which uses one large (16x16) convolution kernel, the deep PEL uses four convolution kernels with kernel size $\{7,3,3,2\}$, stride $\{2,1,1,2\}$, and channel number $\{64,64,64,C\}$. By using small kernel sizes and more convolution kernels, deep PEL helps a vision model to explore the locality inside single patch embedding vector. We mark models with deep PEL as "Hybrid-MS-*+".

Table \ref{tab:exp4} shows comparison between our hybrid models and some of the state-of-the-art models based on CNN, Transformer, or MLP. All listed models are trained on ImageNet-1K. Within the section of our models, we can find that hybrid models achieve better model size-performance trade-off than pure convolution models or Transformer models. The Hybrid-MS-XS achieves 82.4\% top-1 accuracy with 28M parameters, which is higher than Conv-MS-S with 44M parameters and only a little lower than Trans-MS-S with 40M parameters. In addition, the Hybrid-MS-S achieve 83.7\% top-1 accuracy with 63M parameters, which has 0.8 point gain compared with Trans-MS-S. 

The Hybrid-MS-S+ model we proposed achieves 83.9\% top-1 accuracy with 63M parameters. This number is higher than the accuracy achieved by SOTA models Swin-B and CaiT-S36, which have model size of 88M and 68.2M, respectively. The FLOPs of our model is also fewer than these two models. We believe Hybrid-MS-S can be serve as a strong yet simple baseline for future research on architecture design of vision models.

\begin{table}
  \centering
  \begin{tabular}{l|rr|c}
    \toprule
    \multirow{2}{*}{Model}       & \multirow{2}{*}{\#param.}  & \multirow{2}{*}{FLOPs}  & IN-1K  \\
                                 &                            &                         & Top-1 acc. \\
    \hline
    \multicolumn{4}{c}{CNN} \\
    \hline
    RegNetY-4G  \cite{DBLP:conf/cvpr/RadosavovicKGHD20}  & 21M                        & 4.1G                    & 80.0 \\
    RegNetY-8G  \cite{DBLP:conf/cvpr/RadosavovicKGHD20}  & 39M                        & 8.0G                    & 81.7 \\
    RegNetY-16G \cite{DBLP:conf/cvpr/RadosavovicKGHD20}  & 84M                        & 16.0G                   & 82.9 \\
    \hline
    \multicolumn{4}{c}{Transformer} \\
    \hline
    ViT-B/16* \cite{DBLP:conf/iclr/DosovitskiyB0WZ21}    & 86M                        & -                       & 77.9 \\
    DeiT-S \cite{DBLP:conf/icml/TouvronCDMSJ21}          & 22M                        & 4.6G                    & 79.8 \\
    DeiT-B \cite{DBLP:conf/icml/TouvronCDMSJ21}          & 86M                        & 17.5G                   & 81.8 \\
    Swin-T \cite{DBLP:journals/corr/abs-2103-14030}      & 29M                        & 4.5G                    & 81.3 \\
    Swin-S \cite{DBLP:journals/corr/abs-2103-14030}      & 50M                        & 8.7G                    & 83.0 \\
    Swin-B \cite{DBLP:journals/corr/abs-2103-14030}      & 88M                        & 15.4G                   & 83.5 \\

    CaiT-XS24 \cite{DBLP:journals/corr/abs-2103-17239}   & 26.6M                      & 5.4G                    & 81.8 \\
    CaiT-S36  \cite{DBLP:journals/corr/abs-2103-17239}   & 68.2M                      & 13.9G                   & 83.3 \\
    CvT-13 \cite{DBLP:journals/corr/abs-2103-15808}      & 20M                        & 4.5G                    & 81.6 \\
    CvT-21 \cite{DBLP:journals/corr/abs-2103-15808}      & 32M                        & 7.1G                    & 82.5 \\
    \hline
    \multicolumn{4}{c}{MLP} \\
    \hline
    FF-Base \cite{DBLP:journals/corr/abs-2105-02723}     & 62M                        & -                       & 74.9 \\ 
    Mixer-B/16 \cite{DBLP:journals/corr/abs-2105-01601}  & 79M                        & -                       & 76.4 \\
    ResMLP-S24 \cite{DBLP:journals/corr/abs-2105-03404}  & 30M                       & 6.0G                    & 79.4 \\
    ResMLP-B24 \cite{DBLP:journals/corr/abs-2105-03404}  & 45M                       & 23.0G                   & 81.0 \\
    gMLP-S \cite{DBLP:journals/corr/abs-2105-08050}      & 20M                        & 4.5G                    & 79.4 \\
    gMLP-B \cite{DBLP:journals/corr/abs-2105-08050}      & 73M                        & 15.8G                   & 81.6 \\
    \hline
    \multicolumn{4}{c}{Ours} \\
    \hline
    Conv-MS-XS                   & 17M                        & 2.8G                    & 80.1\\
    Conv-MS-S                    & 44M                        & 7.2G                    & 81.6\\
    Trans-MS-XS                  & 14M                        & 3.1G                    & 80.1\\
    Trans-MS-S                   & 40M                        & 7.6G                    & 82.9\\
    \hline
    Hybrid-MS-XS                 & 28M                        & 4.5G                    & 82.4 \\
    Hybrid-MS-XS+                & 28M                        & 5.6G                    & 82.8 \\
    Hybrid-MS-S                  & 63M                        & 11.2G                   & 83.7 \\
    Hybrid-MS-S+                 & 63M                        & 12.3G                   & 83.9 \\

    \bottomrule
  \end{tabular}
  \caption{Comparison of different models on ImageNet-1K classification. Compared models are grouped according to network structures, and our models are listed in the last, Most models are pre-trained with 224x224 images, except ViT-B/16*, which uses 384x384 images.}
  \label{tab:exp4}
\end{table}

\section{Conclusion}

The objective of this work is to understand how the emerging Transformer and MLP structures compare with CNNs in the computer vision domain. We first built a simple and unified framework, called SPACH, that could use CNN, Transformer, or MLP as plug-and-play components. Under the SPACH framework, we discover with a little surprise that all three network structures are similarly competitive in terms of the accuracy-complexity trade-off, although they show distinctive properties when the network scales up. In addition to the analysis of specific network structures, we also investigate two important design choices, namely multi-stage framework and local modeling, which are largely overlooked in previous work. Finally, inspired by the analysis, we propose two hybrid models which achieve SOTA performance on ImageNet-1k classification without bells and whistles.

Our work also raises several questions worth exploring. First, realizing the fact that the performance of MLP-based models is largely affected by over-fitting, is it possible to design a high-performing MLP model that is not subject to over-fitting? Second, current analyses suggest that neither convolution nor Transformer is the optimal structure across all model sizes. What is the best way to fuse these two structures? Last but not least, do better visual models exist beyond the known structures including CNN, Transformer, and MLP?
{\small
\bibliographystyle{ieee_fullname}
\bibliography{egbib}
}

\end{document}